\documentclass[letterpaper, 10 pt, conference]{ieeeconf}  

\IEEEoverridecommandlockouts                              

\usepackage{color}
\definecolor{green}{rgb}{0, 0.5, 0}
\definecolor{orange}{rgb}{0.8, 0.6, 0.2}
\definecolor{red}{rgb}{1.0, 0.0, 0.0}
\definecolor{teal}{rgb}{0.0, 0.4, 0.4}
\definecolor{purple}{rgb}{0.65,0,0.65}
\definecolor{saffron}{rgb}{0.95,0.75,0.2}
\definecolor{turquoise}{rgb}{0.0,0.5,0.5}
\definecolor {mygray}{gray}{.9}

\usepackage[normalem]{ulem}

\usepackage{cite}
\usepackage{amsmath}
\usepackage{amssymb}
\usepackage{amsfonts}

\def\*#1{\mathbf{#1}}

\usepackage{hyperref}

\usepackage{overpic}
\usepackage{graphicx}
\usepackage{caption}
\usepackage{subcaption}

\usepackage{algorithm}
\usepackage{algpseudocode}
\usepackage{pifont}
\usepackage{algorithmicx}

\usepackage{booktabs}
\usepackage{colortbl}
\usepackage{makecell}
\usepackage{multirow}
\usepackage{threeparttable}

\let\labelindent\relax
\usepackage[inline]{enumitem}

\newcolumntype{L}[1]{>{\raggedright\arraybackslash}p{#1}}
\newcolumntype{C}[1]{>{\centering\arraybackslash}p{#1}}
\newcolumntype{R}[1]{>{\raggedleft\arraybackslash}p{#1}}

\usepackage{gensymb}

\title{\LARGE \bf
A Unified BEV Model for Joint Learning of 3D Local Features and Overlap Estimation
\vspace{-3mm}
}

\author{Lin Li$^{1,2}$, Wendong Ding$^{1}$, Yongkun Wen$^{1}$, Yufei Liang$^{2}$, Yong Liu$^{2,*}$ and Guowei Wan$^{1,*}$\vspace{-3mm} 
\thanks{$^{1}$Wendong Ding, Yongkun Wen and Guowei Wan are with Baidu Intelligent Driving Group, Beijing 100094, P. R. China. This work is done when Lin Li is an intern at Baidu.}
\thanks{$^{2}$Lin Li, Yufei Liang and Yong Liu are with the Institute of Cyber-Systems and Control, Zhejiang University, Hangzhou 310027, P. R. China.}
\thanks{$^{*}$Corresponding authors, email: \texttt{\tt\small wanguowei@baidu.com} and \texttt{\tt\small yongliu@iipc.zju.edu.cn.}}
}

\begin{document}

\maketitle
\thispagestyle{empty}
\pagestyle{empty}

\begin{abstract}
	Pairwise point cloud registration is a critical task for many applications, which heavily depends on finding correct correspondences from the two point clouds.
	However, the low overlap between input point clouds causes the registration to fail easily, leading to mistaken overlapping and mismatched correspondences, especially in scenes where non-overlapping regions contain similar structures.
	In this paper, we present a unified bird's-eye view (BEV) model for jointly learning of 3D local features and overlap estimation to fulfill pairwise registration and loop closure.
	Feature description is performed by a sparse UNet-like network based on BEV representation, and 3D keypoints are extracted by a detection head for 2D locations, and a regression head for heights.
	For overlap detection, a cross-attention module is applied for interacting contextual information of input point clouds, followed by a classification head to estimate the overlapping region.
	We evaluate our unified model extensively on the KITTI dataset and Apollo-SouthBay dataset.
	The experiments demonstrate that our method significantly outperforms existing methods on overlap estimation, especially in scenes with small overlaps.
	It also achieves top registration performance on both datasets in terms of translation and rotation errors.
\end{abstract}
\section{INTRODUCTION}
\label{section:intro}

Pairwise point cloud registration aims to align two partially overlapped point clouds, which is a fundamental task in many applications, such as LiDAR SLAM~\cite{Deschaud2018IMLSSLAMSM, ZhangS14LOAM}, LiDAR-based mapping~\cite{Shiratori2015Map, Yang2018Map}, and localization~\cite{Yoneda2014Loc, lu2019l3}.
Another equally important module in the SLAM system is loop closure, which ensures a globally consistent map.
Recent works have made substantial progress in loop closure detection~\cite{pointnetvlad,overlapnet,sc,ssc} and point cloud registration~\cite{d3feat,predator,geometric_transformer}. 
For loop closure detection, it is common practice to encode the entire point cloud into a global descriptor~\cite{pointnetvlad,overlapnet}. 
The advantage of this encoding method is that it is lightweight and convenient for retrieval. 
However, due to the lack of information interaction, this encoding is not robust to occlusions and small overlaps. 
The same problem exists in the field of point cloud registration. 
Some recent point cloud registration works~\cite{predator,xu2021omnet,imlovenet} have begun to focus on small overlapping scenarios. 
However, most of these works are mainly aimed at indoor scenes.
Point cloud registration of outdoor scenes with low overlap is very challenging because the point cloud gets sparser with distance.

\begin{figure}[t]
	\centering
	\includegraphics[width=\linewidth]{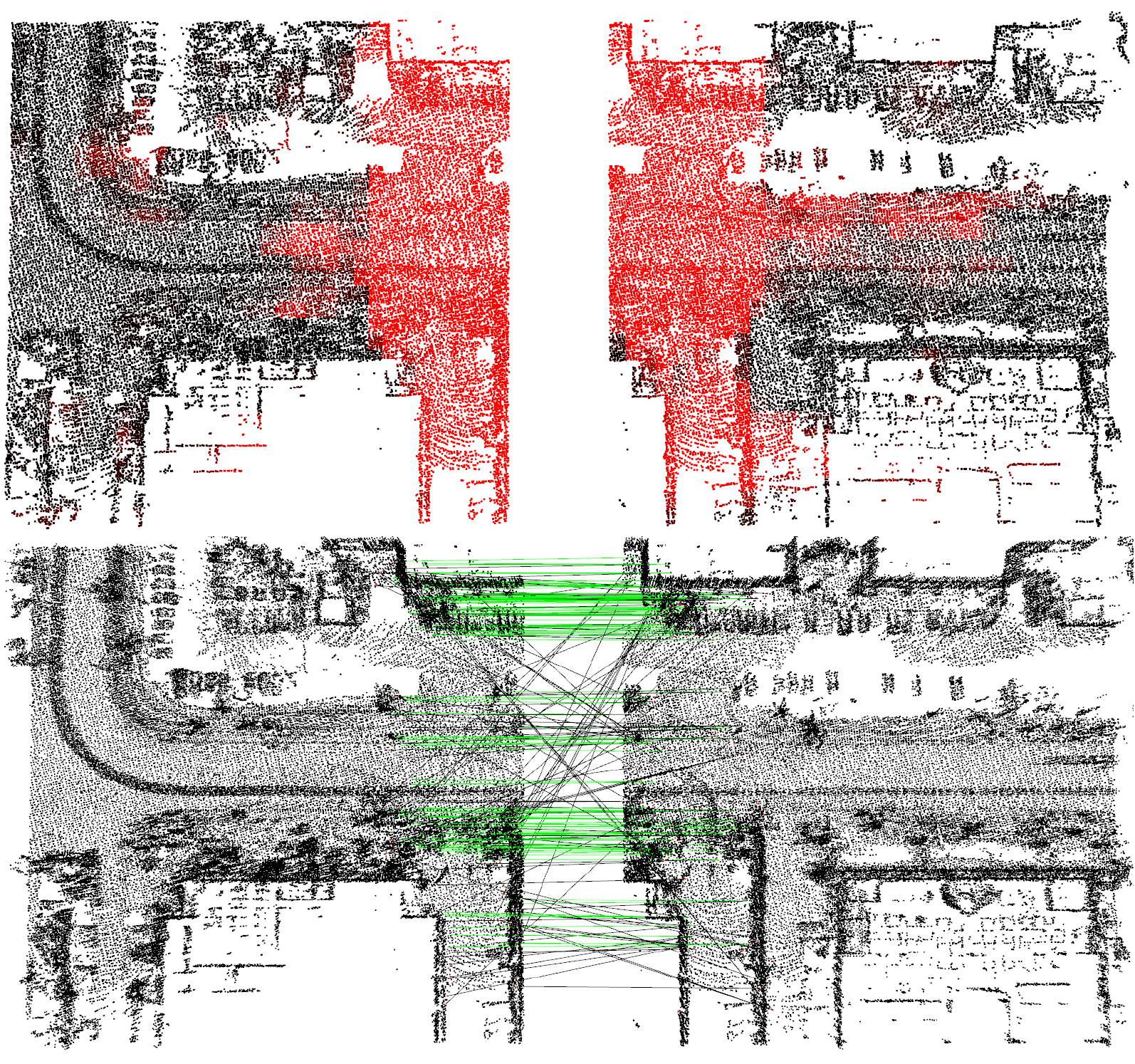}
	\caption{
		\small
		An illustration of our method. 
		The upper part of the figure shows the detected overlapping regions (in red) in an outdoor scene. 
		The lower part shows the correspondences (inliers in green and outliers in black) found by our method.
	}
	\label{fig:demo}
	\vspace{-0.0cm}
\end{figure}

Loop closure detection is inherently related to overlap estimation, and the latter can be considered as a similarity metric.
Intuitively, pairwise registration is directly affected by the overlap of the two point clouds; e.g., the overlap can be utilized to filter out mismatched correspondences.
To go further, in the training stage, the overlap can effectively supervise the contrastive learning of the two input point clouds for feature description and keypoint detection.
Therefore, overlap estimation is important for loop closure and pairwise registration of point clouds.

In this work, we seek to jointly learn overlap estimation and 3D local features in a unified BEV model.
BEV form is a compact and natural representation of point clouds in outdoor scenes, which are usually collected using LiDAR sensors mounted on vehicles.
We represent input point clouds as multi-layer BEVs and apply a UNet-like network to extract multi-scale features. 
We detect keypoints on the 2D BEV representation and extract local descriptors.
For 6-DoF registration, we obtain the height of each 2D BEV cell in a regression way.
We adopt a cross-attention module to interact with the input point clouds and then perform overlapping region classification on the 2D BEV plane.
For pairwise registration, we only detect corresponding keypoints within the overlapping area.
For loop closure detection, we take the area of the overlapping region as a measure of their similarity.
Fig.~\ref{fig:demo} is an illustration of our method.

To summarize, our main contributions are:

\begin{itemize}
	\item A joint learning framework for overlap estimation and 3D local features, which effectively fulfills loop closure and  6-DoF registration in urban scenes.
	\item A novel overlap estimation method that fully interacts with pairwise information, yielding high precision and recall under low overlap scenes. 
	\item Based on BEV, the separation of 2D keypoint detection on the BEV plane and height regression makes it an efficient and practical 3D keypoint detection method.
	\item Rigorous tests and detailed ablations on the KITTI\cite{geiger2012we} and Apollo-Southbay\cite{lu2019l3} datasets to comprehensively verify the effectiveness of the proposed method. 
\end{itemize}

\section{RELATED WORK}
\subsection{LiDAR-based Loop Closure Detection}
As a hot research field of SLAM, numerous researchers have studied LiDAR-based loop closure detection, and many excellent works have been proposed~\cite{isc,ssc,seed,iris,sc_plusplus,pointnetvlad,lpdnet,spoxelnet,ndttransformer,fusionvlad,overlaptransformer}.

One of the common solutions is encoding the input point cloud into a global descriptor\cite{m2dp,delight} as a 1D vector or 2D matrix and comparing their similarity to find the loop closures. 
Scan Context~\cite{sc} globally describes the point cloud as a bird's-eye view (BEV) with height information in polar coordinates, which makes the descriptor robust to rotation and has good generalization.
Extension works~\cite{isc,ssc,seed,iris,sc_plusplus} encode additional semantic~\cite{ssc} and intensity~\cite{isc} information further to improve the detection performance.

Recently learning-based methods have shown impressive results.
Some works~\cite{pointnetvlad,lpdnet,spoxelnet,ndttransformer,fusionvlad,overlaptransformer} extract local features with a deep network and aggregate them to a global descriptor with NetVLAD~\cite{netvlad} or other context gating techniques\cite{rinet,disco,minkloc3d}.
Another common practice is to segment the point cloud into objects as local features and then match them directly~\cite{segmatch,segmap} or by a graph~\cite{sgpr,gosmatch}.

The above methods only encode features from its single input point cloud and do not know about the correlated information from the counterpart one.
Predator~\cite{predator} fuses the feature maps from two stream networks and implicitly encodes the overlapping context with designated supervision loss to handle the low overlap registration problem.
Unlike Predator, overlap estimation is an explicit network design in our method, rather than supervision only, which is more conducive to obtaining desired results.
Furthermore, we use the deepest feature maps that contain contextual information for interacting instead of at the point level, which makes our method more robust and is validated in our experiments.

\subsection{Deep Point Cloud Registration}
Point cloud registration is also a widely studied topic in SLAM research society, in which deep learning methods demonstrate promising results when solving challenge cases, e.g., bad initialization or low overlap. 
Some of these methods are based on directly inferring correspondences, in which key points are extracted and described on local patches~\cite{ppffoldnet,perfectmatch,3dmatch}, then the accurate transformation is obtained with robust pose estimation, e.g., RANSAC~\cite{lu2019deepvcp} or weighted Procrustes~\cite{choy2020deep}.
In order to learn local features more effectively, point convolution backbones~\cite{choy20194d,kpconv} are adopted to extract dense features in a single forward process~\cite{fcgf,d3feat,predator}.

Instead of directly finding correspondences, some other methods estimate the transformation in an end-to-end manner.
Some of them~\cite{fu2021robust,wang2019prnet,yew2020rpm} build soft correspondences by learning patch features and use a differentiable weighted SVD module to compute the transformation. 
Others~\cite{pointnetlk,huang2020feature,xu2021omnet} directly use the extracted features to regress the transformation.

These learning-based methods give competitive results, but the performance drops drastically in small overlap scenes.
This problem already draws the attention of many researchers, as \cite{predator,geometric_transformer,imlovenet} tried in indoor scenes.
Our method only detects keypoints in the estimated overlap, thus avoiding wrong matches between non-overlapping regions.

\section{METHODOLOGY}
\begin{figure*}[t]
	\centering
	\includegraphics[width=.8\linewidth]{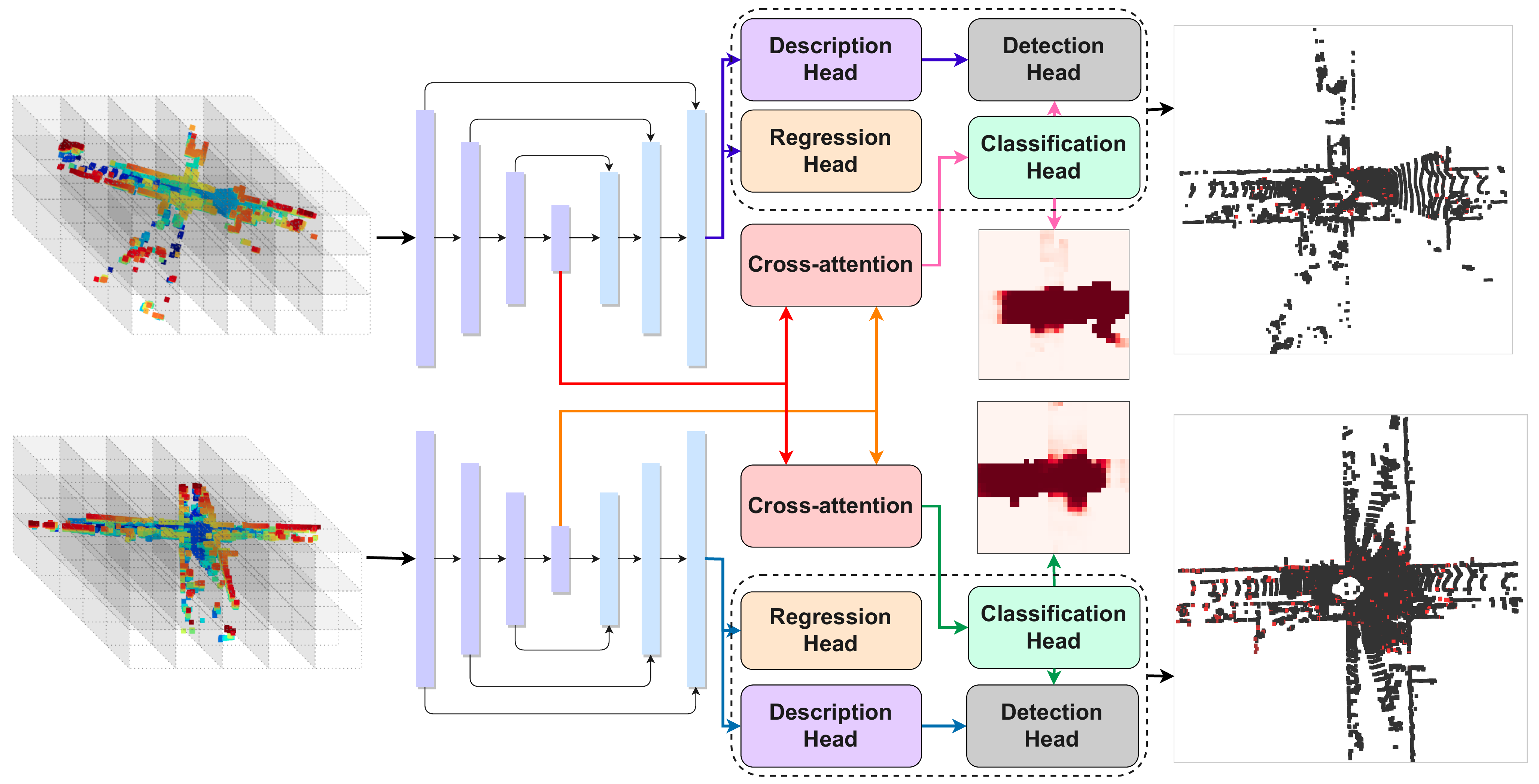}
	\caption{
		\small
		The architecture of the proposed unified BEV model for 3D local features and overlap estimation. We take multi-layer BEV representations as 
		the input of the 2D UNet backbone. The feature maps of the last layers of the encoder and decoder are used for overlapping 
		region detection and local feature extraction, respectively.
	}
	\label{fig:system}
	\vspace{-0.5cm}
\end{figure*}

In this section, we describe the architecture of the proposed unified BEV model for 3D local features and overlap estimation in detail, as shown in Fig.~\ref{fig:system}. 
Inspired by DiSCO\cite{disco}, the two input point clouds are first converted to multi-layer BEV representations and then fed into a 2D U-Net backbone to extract multi-scale features.
The deepest feature maps of the two input point clouds interact with each other via a cross-attention module to measure their relevance, after which a classification head is used to calculate two overlap score maps.
Meanwhile, the multi-scale features are fed into a description head, a detection head, and a regression head to obtain feature descriptors and 3D keypoints, respectively.


\subsection{Dense Feature Description}

We first divide the input point clouds $\mathcal{P}$ and $\mathcal{Q}$ into $H\times W\times C$ grids, in which each voxel is set to 0 or 1 depending on its occupancy.
By treating each pillar of the grid as a $C$-dimension channel, the point clouds $\mathcal{P}$ and $\mathcal{Q}$ are then converted into BEV representations, denoted as $B_P \in \mathbb{R}^{H\times W\times C}$ and $B_Q \in \mathbb{R}^{H\times W\times C}$. 



\textbf{2D UNet Backbone.}
Instead of using 3D (sparse) convolutions or point convolutions, we directly apply 2D sparse convolutions on BEV representations to extract deep features. Concretely, we use a 2D UNet-like structure with skip connections and residual blocks in the encoder and decoder. Considering the sparsity of the inputs, 2D sparse convolutions can be used to speed up. After performing 2D convolutions on BEV representations, we can obtain the following multi-scale feature maps:
\begin{equation}
	E_t^1, \dots, E_t^s = \mathcal{E}(B_t),~t\in \{\mathcal{P},~\mathcal{Q}\}
\end{equation}
\begin{equation}
	F_t^{s-1}, \cdots, F_t^1 = \mathcal{F}(E_t),~t\in \{\mathcal{P},~\mathcal{Q}\},
\end{equation}
where $\mathcal{E}$ and $\mathcal{F}$ represent the encoder and decoder of the backbone, respectively.



\textbf{Description Head.}
The description head extracts feature descriptors $D_t$ from the feature map $F_t^1$, which consists of a $1 \times 1$ convolution and normalization layer as follows:
\begin{equation}
	D_t = \mathrm{Norm_{L_2}}(\mathrm{Conv}_{1 \times 1}(F_t^1)),~t\in \{\mathcal{P},~\mathcal{Q}\},
\end{equation}
where $\mathrm{Norm_{L_2}}$ is the $\mathrm{L_2}$ normalization operation across feature channels.

\subsection{Dense Keypoint Detection}
D3Feat \cite{d3feat} detects 3D keypoints of point cloud based on the loal maximum of the channel and spatial dimensions of the point features. Our keypoint detection also adopts a similar way, except that we detect 2D keypoints \{($x$, $y$)\} on the BEV feature descriptors $D_t$ and regress their heights \{$z$\} to form the 3D keypoints \{($x$, $y$, $z$)\}, which makes our approach a more efficient implementation.


\textbf{Detection Head.} 
The spatial saliency of each pixel in $D_t$ is evaluated in its local neighborhood. Thanks to the regularity of BEV representation, the neighborhood of each pixel simply consists of the pixels within the square centered around it, thus avoiding the heavy operation of kdtree search in D3Feat \cite{d3feat}. The spatial saliency score of each pixel $p_{ij}$ is defined as

\begin{equation} \label{spatial_score}
	\alpha_{ij}^k = \ln\left(1+\exp\left(D_{ij}^k - \frac{1}{|\mathcal{N}_{ij}|}\sum_{(i'j') \in \mathcal{N}_{ij}} D_{i'j'}^k\right)\right),
\end{equation}
where $k=1,...,C$, and $\mathcal{N}_{ij}$ represents the non-empty neighboring pixels of $p_{ij}$.

For each pixel $p_{ij}$, there will be at most $s \times s$ non-empty pixels in its neighborhood, where $s$ represents the length of the square. In this way, for Equ. ~\ref{spatial_score}, we can use the AvgPool operation to achieve an efficient implementation as follows. 

\begin{equation} \label{spatial_score2}
	\alpha_{ij}^k = \ln\left(1+\exp\left(D_{ij}^k - \frac{s^2 \times \mathrm{AvgPool}(D)_{ij}^k}{s^2 \times \mathrm{AvgPool}(B^*)_{ij}}\right)\right),
\end{equation}
where $s$ is the window size of the average pooling, and $s^2 \times \mathrm{AvgPool}(D)_{ij}^k$ is the sum of $k$-th channel values of neighborhood, $B^*=\max_k(B^k)$ is the channel max value representing the occupancy of pillars, and $s^2 \times \mathrm{AvgPool}(B^*)_{ij}$ represents the number of the non-empty neighboring pixels. The $s^2$ in the numerator and denominator can be eliminated. In addition, a sparse AvgPool operation can be directly used to calculate the average of non-empty pixels in Equ.~\ref{spatial_score}, thus replacing Equ.~\ref{spatial_score2}.

The channel max score is computed as

\begin{equation}
	\beta_{ij}^k = \frac{D_{ij}^k}{\max_c D_{ij}^c}.
\end{equation}

Both the spatial and channel scores are considered for computing the final detection score:
\begin{equation}
	s_{ij}=\max_k (\alpha_{ij}^k \beta_{ij}^k).
\end{equation}

\textbf{Regression Head.} 
The 2D salient keypoints can be detected in BEV representation with $s_{ij}$, but the heights still need to be recovered.
Here we apply a regression head to predict a weight vector $W_{ij}\in [0,1]^{C\times 1}$ for each pillar $B_{ij}$. 
Let $H_{ij}\in \mathbb{R}^{C\times 1}$ denote the heights of voxels in a pillar, then the height of the keypoint in $B_{ij}$ is predicted  by a convolutional layer and a sigmoid layer as
\begin{equation}
	W=\text{Sigmoid}(\mathrm{Conv}_{3\times3}(F_t^1))
\end{equation}
\begin{equation}
	z_{ij}=W_{ij}^T*H_{ij}.
\end{equation}

Finally, we obtain a regressed height $z$ for each non-empty pillar and denote the regressed point clouds as $\mathcal{P}'$ and $\mathcal{Q}'$.

\subsection{Overlapping Region Classification}\label{method:overlap}
Cross attention has demonstrated its effectiveness in interacting information \cite{predator,geometric_transformer} and detecting overlap regions \cite{imlovenet} from encoded feature maps.
Similar to ImLoveNet \cite{imlovenet}, we adopt cross attention on two feature maps of the input point clouds to learn relevant information, followed by a classification head to solve the overlap as learning a similarity score. Different from \cite{imlovenet}, we only use the deepest feature maps for overlap estimation 
because the deepest feature maps contain richer context information and are easier to learn robust correlations. 




\textbf{Cross Attention.} 
The cross attention module takes the deepest feature maps, $E_P^s \in \mathbb{R}^{H_s \times W_s \times C_s}$ and $E_Q^s\in \mathbb{R}^{H_s \times W_s \times C_s}$, to generate two relevant feature maps, $M_P \in \mathbb{R}^{H_s \times W_s \times C_s}$ and $M_Q \in \mathbb{R}^{H_s \times W_s \times C_s}$, in a bilateral way. Specific details are as follows.
\begin{equation}
	\begin{aligned}
		M_P =& E_P^s+\mathrm{MLP}(\mathrm{cat}(E_P^s, \mathrm{att}(E_P^s, E_Q^s, E_Q^s)))\\
		M_Q =& E_Q^s+\mathrm{MLP}(\mathrm{cat}(E_Q^s, \mathrm{att}(E_Q^s, E_P^s, E_P^s))),
	\end{aligned}
\end{equation}
where $\mathrm{MLP}(\cdot)$ denotes a three-layer fully connected network, and $\mathrm{att}$ is the attention model, the detailed description can be referred from~\cite{imlovenet}.

\textbf{Classification Head.} 
With the correlated feature maps $M_P$ and $M_Q$ from the cross attention module, we apply a binary classification to predict the overlap score maps, $\gamma_P \in [0, 1]^{H_s \times W_s}$ and $\gamma_Q \in [0, 1]^{H_s \times W_s}$, of $\mathcal{P}$ and $\mathcal{Q}$ as
\begin{equation}
	\gamma_t = \mathrm{Sigmoid}\left(\mathrm{Conv_{3\times3}}\left(\mathrm{ReLU}\left(\mathrm{Conv_{3\times 3}\left(M_t\right)}\right)\right)\right),
\end{equation}
where two $3\times3$ convolution layers, a sigmoid layer, and a ReLU are used.

\textbf{Similarity Score.} 
By counting the overlapped regions, we can obtain a similarity metric as
\begin{equation}
	\tau = \frac{1}{2}\left(\frac{\sum \gamma_P}{V_P} + \frac{\sum \gamma_Q}{V_Q}\right),
\end{equation}
where $V_P$ and $V_Q$ denote the number of occupied pixels of $M_P$ and $M_Q$. In subsequent experiments~\ref{exp:overlap}, we will demonstrate that this similarity metric can be used for the loop closure detection task.

\subsection{Loss Function}
\label{section:loss}

To train the network in an end-to-end manner, we utilize multi-task loss functions for jointly optimizing the feature description, keypoint detection, height regression, and overlap region classification.


\textbf{Description Loss.} 
Following~\cite{predator}, we take the circle loss~\cite{circle_loss} to learn discriminative descriptors. 
We perform random sampling to balance the number of positive and negative samples. 
The positive samples $\Omega_p$ are selected correspondences, where the set of correspondences is defined as points in $\mathcal{Q}'$ that lie within a radius around point $i$ in $\mathcal{P}'$.
The negative samples $\Omega_n$ are formed from points of $\mathcal{Q}'$ outside a larger radius of the point $i$.
This loss function can be expressed by:
\begin{equation}\label{equ:desc_loss}
	L_{desc}^P=\frac{1}{N}\sum_{i=1}^{N}\mathrm{ln}\left(1+\sum_{j\in\Omega_p}e^{\theta_p^j(d_i^j-\Delta_p)}\cdot \sum_{k\in\Omega_n}e^{\theta_n^k(\Delta_n-d_i^k)}\right),
\end{equation}
where $\Delta_p$ and $\Delta_n$ are positive and negative margins, $d_i^j$ and $d_i^k$ are feature distance of 
positive samples and negative samples, 
$\theta_p^j$ and $\theta_n^k$ are the positive and negative weights, computed for each sample individually with $\theta^j_p=\gamma(d_i^j-\Delta_p)$ and $\theta^k_n=\gamma(\Delta_n-d_i^k)$. 
We recommend referring to the original paper~\cite{circle_loss} for details. 
Through the same process, we can get the reverse loss $L_{desc}^Q$, and the total loss $L_{desc}$ is the average of $L_{desc}^P$ and $L_{desc}^Q$. 

\textbf{Detection Loss.} 
The detection loss aims to encourage the easily matchable correspondences to have higher keypoint detection scores than the correspondences which are hard to match as
\begin{equation}
		L_{det}=\frac{1}{N}\sum_i\left(d^\text{pos}_i-d^\text{neg}_i\right)\left(s_{P_i}+s_{Q_i}\right),
\end{equation}
where $(P_i, Q_i)$ are correspondences of $\mathcal{P}'$ and $\mathcal{Q}'$, $s_{P_i}$ and $s_{Q_i}$ denote their saliency scores, 
$d^\text{pos}_i$ is the feature distance between positive samples, $d^\text{neg}_i$ represents the feature distance between the hardest negative samples.

\textbf{Regression Loss.} 
For the input point cloud $\mathcal{P}$, to recover the heights of keypoints, we use both the origin point cloud $\mathcal{P}$ and its counterpart $\mathcal{Q}$ to supervise the recovered heights as
\begin{equation}
	L_{reg}^P=\frac{1}{N}\sum_i\left(\| z_{P'_i}-z_{P_j}\|+\| z_{P'_i}-z_{Q'^T_i}\|\right),
\end{equation}
where $z_{P'_i}$ is the predicted height of point $P'_i$ in regressed point cloud $\mathcal{P}'$, $z_{P_j}$ is the height of closest point $P_j$ to $P'_i$ in point cloud $\mathcal{P}$, and $Q'^T$ is the point cloud $\mathcal{Q}'$ transformed into frame of $\mathcal{P}'$, $z_{Q'^T_i}$ is the predicted height of corresponding point of $P'_i$ in the $\mathcal{Q}'^T$.

\textbf{Classification Loss.}
A binary cross entropy is used in the classification loss as
\begin{equation}
	L_{bce}=\mathrm{BCE}(\gamma_P,l_P)+\mathrm{BCE}(\gamma_Q,l_Q),
\end{equation}
where $\mathrm{BCE}$ denotes the binary cross entropy, $l_P\in[0,1]^{H_s\times W_s}$ and $l_Q\in[0,1]^{H_s\times W_s}$ are ground-truth labels.
In addition, we found that additional supervision of strengthening contrastive distance on the deepest feature map is helpful for network convergence.
Therefore we construct another circle loss $L_{sg}$ similar to Equ.~\ref{equ:desc_loss} on the deepest feature map and forms the classification loss together with BCE loss $L_{bce}$.


\section{EXPERIMENTS}

\begin{figure*}[t]
	\centering
	\includegraphics[width=.9\linewidth]{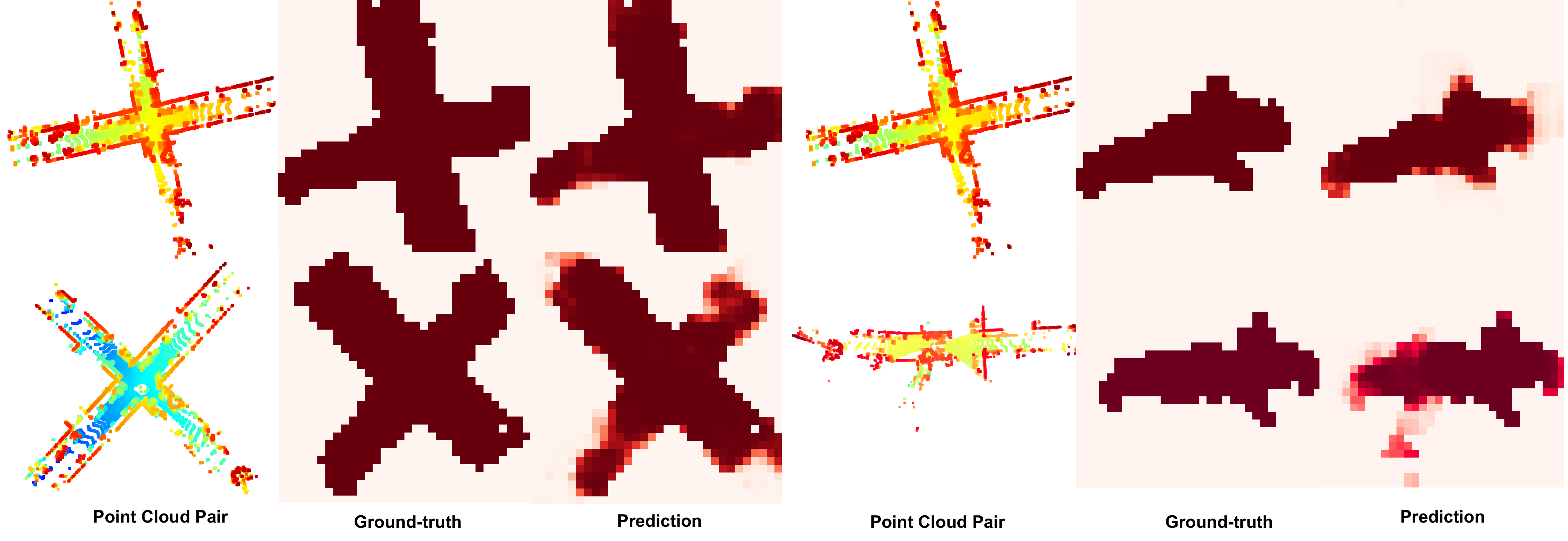}
	\caption{
		\small
		We show the results of overlapping region detection on two pairs of point clouds with different distances. 
		The first and fourth columns are the original point clouds, the second and fifth columns are the actual overlapping regions, and the 
        third and sixth columns are the predicted overlapping regions.
	}
	\label{fig:overlap}
\end{figure*}

\subsection{Datasets and Implementation Details}
\label{dataset}
Our method has been extensively tested in real-world urban scenarios, primarily using two public datasets, the KITTI dataset~\cite{geiger2012we}, and the Apollo-SouthBay dataset~\cite{lu2019l3, lu2019deepvcp}.
Code is available at \href{https://github.com/lilin-hitcrt/BEVNet}{https://github.com/lilin-hitcrt/BEVNet}.

\textbf{KITTI Odometry Dataset.}
The KITTI odometry dataset collected point clouds captured with a Velodyne HDL64 LiDAR. 
It contains a total of 22 sequences, of which only the first 11 have ground-truth pose annotations. 
We use the last 11 sequences to train our network and use the poses provided by semantic KITTI~\cite{semantic_KITTI} for supervision. 
We validate the performance of our method in detecting loop closures on six sequences (00, 02, 05, 06, 07, 08). 
Like other methods, we verified the performance of point cloud registration on 08-10 sequences.

\textbf{Apollo-SouthBay Dataset.}
The Apollo-SouthBay dataset collected point clouds using the same model of LiDAR as the KITTI odometry dataset, but in the San Francisco Bay area, United States. 
Similar to KITTI, it covers various scenarios, including residential areas, urban downtown areas, and highways. 
Our model is trained on sequence Columbia-Park and tested on sequence Sunnyvale-Big-Loop to demonstrate the method's performance on small overlap loop closure detection and point cloud registration. 
Considering points are sparse in the region far from the LiDAR center in a single frame, we stitch point clouds from several consecutive frames into a \textit{submap}.
Additionally, the submap is cropped within 100m$\times$ 100m and voxelized into 50cm voxels for use.

\textbf{Implementation Details.}
The BEV representation is formed into the shape of $256\times256\times32$.
The backbone is configured with four layers in the encoder and three layers in the decoder, so the deepest feature, $E^4$, has the shape of $32\times32\times512$. 
In constructing the training data, point cloud pairs with distances ranging from 0 to 80 meters are used.
We adopt spconv~\cite{yan2018second} to implement our backbone.
Our code is based on PyTorch using the Adam optimizer with a learning rate of $10^{-4}$.

\begin{table}[tbp]\footnotesize
    \caption{\centering Overlap classification results on KITTI dataset}\vspace{-3mm}
    \label{table:kitti_overlap}
    \begin{center}
    \begin{threeparttable}
        {
    \begin{tabular}{c| c c| c c| c c}
    \hline
    \multirow{2}{*}{Distance(m)}&\multicolumn{2}{c}{IOU(\%)}&\multicolumn{2}{c}{Precision(\%)}&\multicolumn{2}{c}{Recall(\%)}\\
    ~&OP~\cite{predator}&Ours&OP~\cite{predator}&Ours&OP~\cite{predator}&Ours\\
    \hline
    10&73.5&\textbf{88.8}&91.5&\textbf{99.0}&78.9&\textbf{89.5}\\
    20&65.9&\textbf{85.9}&82.4&\textbf{97.2}&76.6&\textbf{87.9}\\
    30&56.3&\textbf{83.2}&70.0&\textbf{94.9}&74.3&\textbf{86.6}\\
    40&42.5&\textbf{78.5}&52.8&\textbf{91.0}&69.6&\textbf{84.1}\\
    50&29.5&\textbf{74.1}&35.5&\textbf{85.3}&67.1&\textbf{83.2}\\
    60&15.9&\textbf{61.5}&18.1&\textbf{72.5}&63.5&\textbf{76.6}\\
    Mean&47.3&\textbf{78.6}&58.4&\textbf{90.0}&71.7&\textbf{84.7}\\
    \hline
    \end{tabular}
    }
    \begin{tablenotes} 
        \footnotesize
        \item The best scores are marked in bold.
     \end{tablenotes}
    \end{threeparttable}
    \end{center}
	\vspace{-0.4cm}
    \end{table}

\subsection{Overlap Estimation}\label{exp:overlap}
\textbf{Overlapping Region Detection.} 
As shown in Table.~\ref{table:kitti_overlap}, overlapping region evaluation metrics include intersection over union (IOU), classification precision and recall.
On sequences 08-10 of the KITTI odometry dataset, different distances between pairs ranging from 10m to 60m are used to verify the influence of overlap size on our model's performance.
The comparison method~\cite{predator}, marked as OP, was retrained with the same distance configuration.
The result shows that our method can effectively detect overlapping regions with average IOU, precision, and recall of 78.6\%, 90.0\%, and 84.7\%, respectively.
As the distance between pairs is going large, the performance of Predator drops drastically, while our method shows better results in each distance range with only a slight fall, and still above 0.7 even under 60m distance.
One of the causes is that overlap classification on the raw, unorganized point cloud in Predator is more difficult than in our method.  
The ablation studies \ref{exp:ablation} have shown support for this view.
Two pairs of point clouds and their ground truth and predicted overlapping regions are visualized in Fig.~\ref{fig:overlap}.

\begin{table}[tbp]\footnotesize
    \caption{\centering Loop closure detection results on KITTI dataset}\vspace{-3mm}
    \label{table:kitti_loop}
    \begin{center}
    \begin{threeparttable}
        {
    \begin{tabular}{c| c c c c c c c c}
    \hline
    Methods & 00 & 02 & 05 & 06 & 07 &08&Mean\\ 
    \hline
    OT~\cite{overlaptransformer} & 89.3 & 85.2 & 92.8 & \textbf{100.0} & 86.3 & 71.8 & 87.6\\
    DS~\cite{disco} & 90.6 & 86.6 & 90.7 & 99.6 & 91.9 & 88.8 & 91.4\\
    LC~\cite{locus} & 92.5 & - & 91.0 & 98.2 & 92.5 & 90.7&93.0\\
    RI~\cite{rinet} & \textbf{97.5} & 92.7 & 91.2 & \textbf{100.0} & 89.4 & \textbf{97.5}&94.7\\
    Ours & 97.0 & \textbf{94.5} & \textbf{97.2} & 98.9 & \textbf{95.7} & 96.6 & \textbf{96.7}\\
    \hline
    \end{tabular}
    }
    \begin{tablenotes} 
        \footnotesize
        \item Average Recall@1. The best scores are marked in bold.~\cite{locus} failed in sequence 02.
     \end{tablenotes}
    \end{threeparttable}
    \end{center}
    \end{table}

\textbf{Loop Closure Detection.}
To make the model robust to loop closure detection, the issues of occlusion and small overlaps need to be addressed.
To compare with the existing methods, Recall@1 in~\cite{disco} is used as the evaluation metric.
The best matching result for each query frame is inferred among the neighboring frames around the query, excluding 100 consecutive frames near the query.
An inference is considered correct when its distance from the query is less than 10m.
As shown in Table.~\ref{table:kitti_loop}, our method outperforms the existing methods on most sequences. 
We can conclude that estimating the correct overlapping regions is the key factor in making our method stand out.
Most methods could successfully detect loops in large overlapping scenes, while only our method survives with small overlaps, as illustrated in the right part of Fig.~\ref{fig:overlap}.

\begin{table}[tbp]\footnotesize
    \caption{\centering Loop closure detection results on Apollo-SouthBay dataset}\vspace{-3mm}
    \label{table:apollo_loop}
    \begin{center}
    \begin{threeparttable}
        {
    \begin{tabular}{c| c c c c c c c}
    \hline
    Methods & 0-10m & 10-20m & 20-30m & 30-40m & 40-50m &50-60m\\ 
    \hline
    OT~\cite{overlaptransformer} & 86.3 & 34.3 & 15.9 & 15.0 & 10.8 & 9.7\\
    DS~\cite{disco} & 90.8 & 43.8 & 12.5 & 13.7 & 7.0 & 7.9\\
    Ours & \textbf{97.9} & \textbf{85.7} & \textbf{62.5} & \textbf{66.0} & \textbf{57.3} & \textbf{57.1}\\
    \hline
    \end{tabular}
    }
    \begin{tablenotes} 
        \footnotesize
        \item Average Recall@1. The best scores are marked in bold.
     \end{tablenotes}
    \end{threeparttable}
    \end{center}
    \end{table}

Loop closure detection is also tested on the Apollo-Southbay dataset. 
The tested pairs are sampled at various distances with 10m intervals. 
As shown in Tab.~\ref{table:apollo_loop}, our method outperforms others at all distance settings.
Since the Sunnyvale-Big-Loop sequence contains some quite different scenes which are not included in the training data, our method still works well, which demonstrates the better generalization capacity of our method.

\subsection{Point Cloud Registration}\label{exp:registration}
\begin{table}[tbp]\footnotesize
    \caption{\centering Registration results on KITTI dataset}\vspace{-3mm}
    \label{table:kitti_registration}
    \begin{center}
    \begin{threeparttable}
        {
    \begin{tabular}{c|c c c c}
    \hline
    Methods & RTE(cm) & RRE($\circ$) & RR(\%)\\ 
    \hline
    3DFeat~\cite{yew20183dfeat} & 25.9 & 0.25 & 96.0\\
    FCGF~\cite{fcgf} & 9.5 & 0.30 & 96.6\\
    D3Feat~\cite{d3feat} & 7.2& 0.30 & 99.8\\
    SpinNet~\cite{spinnet} & 9.9 & 0.47 & 99.1\\
    Predator~\cite{predator}& 6.8&0.27&99.8\\
    COFiNet~\cite{cofinet}&8.2&0.41&99.8\\
    GeoTransformer~\cite{geometric_transformer}&6.8&0.24&99.8\\
    Ours & 7.5 & 0.26 & 99.8\\
    \hline
    \end{tabular}
    }
    \end{threeparttable}
    \end{center}
	\vspace{-0.2cm}
    \end{table}

\begin{table}[tbp]\footnotesize
    \caption{\centering Registration results at different distances}\vspace{-3mm}
    \label{table:apollo_registration}
    \begin{center}
    \begin{threeparttable}
        {
    \begin{tabular}{c| c c c| c c}
    \hline
    \multirow{2}{*}{Distance(m)}&\multicolumn{3}{c}{KITTI}&\multicolumn{2}{c}{Apollo}\\
    ~&OP~\cite{predator}&O.w.o.O&O.w.O&O.w.o.O&O.w.O\\
    \hline
    10&97.1&\textbf{99.6}&\textbf{99.6}&99.5&\textbf{100.0}\\
    20&95.4&96.8&\textbf{98.2}&99.8&\textbf{99.8}\\
    30&80.5&87.0&\textbf{96.2}&99.3&\textbf{99.7}\\
    40&51.1&59.9&\textbf{86.9}&97.7&\textbf{98.8}\\
    50&24.5&33.0&\textbf{67.9}&93.0&\textbf{97.2}\\
    60&9.4&11.8&\textbf{47.1}&88.2&\textbf{94.5}\\
    70&-&-&-&74.2&\textbf{89.7}\\
    80&-&-&-&47.4&\textbf{76.7}\\
    \hline
    \end{tabular}
    }
    \begin{tablenotes} 
        \footnotesize
        \item Registration Recall (\%). The best scores are marked in bold.
     \end{tablenotes}
    \end{threeparttable}
    \end{center}
	\vspace{-0.5cm}
    \end{table}

For pairwise registration, state-of-the-art methods~\cite{yew20183dfeat,fcgf,d3feat,spinnet,predator,cofinet,geometric_transformer} are compared at a 10m distance setting, and all keypoints are used.
We use RANSAC with 50,000 max iterations to estimate the transformation following \cite{d3feat}.
As shown in Tab.~\ref{table:kitti_registration}, the relative translation error (RTE),  relative rotation error (RRE), and registration recall (RR)\cite{d3feat} of our method are $7.5$ cm, $2^\circ$, and $99.8\%$, respectively. 
The comparable accuracy our method achieved illustrates the effectiveness of our feature description and keypoint detection on BEV representations.

We conduct experiments with various distance settings to address the low overlap cases.
With registration recall defined as $\text{RTE}<2$m and $\text{RRE}<5^\circ$, Predator\cite{predator} and our method with/without overlap estimation detection (O.w.O/O.w.o.O) are compared on the KITTI and Apollo-Southbay datasets.
Up to 250 keypoints per point cloud are used.
As shown in Tab.~\ref{table:apollo_registration}, O.w.o.O achieves comparable results to OP \cite{predator} on KITTI, and O.w.O always keeps top performance, thus illustrating the usefulness of overlap estimation in small overlapping scenes.
The comparison of our method shows the same conclusion on the Apollo-Southbay dataset, O.w.O/O.w.o.O methods give 77\%/47\% RR at 80m, respectively.
The grouped comparison demonstrates that the overlap is crucial for registration:
\textit{A better overlap estimation makes better registration.}

\subsection{Ablation Study}\label{exp:ablation}
\begin{table}[t]\footnotesize
    \caption{\centering Ablation study on registration}\vspace{-3mm}
    \label{table:ablation_loss}
    \begin{center}
    \begin{threeparttable}
        {
    \begin{tabular}{l| c c c}
    \hline
    \multirow{2}{*}{Loss} & \multicolumn{3}{c}{Registration Metrics}\\
	&RR&RTE&RRE\\
    \hline
    $L_{desc}$&1.46&119.4&3.97\\
    $L_{desc}+L_{reg}$&40.1&94.1&2.33\\
    $L_{desc}+L_{reg}+L_{det}$&59.9&63.2&1.77\\
    $L_{desc}+L_{reg}+L_{det}+L_{bce}+L_{sg}$&\textbf{86.9}&\textbf{57.0}&\textbf{1.63}\\
    \hline
    \multirow{2}{*}{Loss} &\multicolumn{3}{c}{Overlap Metrics}\\
	 &OI&OP&OR\\
    \hline
    $L_{desc}+L_{reg}+L_{det}+L_{bce}$&60.2&74.5&75.0\\
    $L_{desc}+L_{reg}+L_{det}+L_{bce}+L_{sg}$&\textbf{78.6}&\textbf{90.0}&\textbf{84.7}\\
    \hline
    \end{tabular}
    }
     \begin{tablenotes} 
         \footnotesize
         \item OP: overlap estimation precision. OR: overlap estimation recall.
         \item OI: overlap estimation IOU.
         \item The best scores are marked in bold.
      \end{tablenotes}
    \end{threeparttable}
    \end{center}
    \end{table}

We ablate the loss functions for registration and overlap estimation tasks on the KITTI dataset.
The ablation of registration uses a 40m distance setting for the point cloud pair and 250 keypoints for each point cloud, while the ablation of overlap estimation is the same as in Section~\ref{exp:overlap}.
As shown in Tab.~\ref{table:ablation_loss}, height information supervised by $L_{reg}$ is indispensable for registration recall.
The loss $L_{bce}+L_{sg}$ for overlap estimation improves the registration accuracy significantly (recall+27\%).
The detection loss $L_{det}$ helping to select discriminate key points also boosts recall with +20\%, RTE with -30cm. 
In addition, the loss term $L_{sg}$ has a remarkable influence on overlap estimation (precision+15\%).
\begin{table}[tbp]\footnotesize
    \caption{\centering Ablation study on overlap estimation}\vspace{-3mm}
    \label{table:ablation_overlap}
    \begin{center}
    \begin{threeparttable}
        {
    \begin{tabular}{c|c c c c}
    \hline
    Feature map&Size&IOU (\%)&Precision (\%)&Recall (\%)\\
    \hline
    $F^1$&$256\times256$&50.3&62.2&74.2\\
    $F^2$&$128\times128$&64.9&78.2&78.7\\
    $F^3$&$64\times64$&73.8&85.2&83.2\\
    $E^4$&$32\times32$&\textbf{78.6}&\textbf{90.0}&\textbf{84.7}\\
    \hline
    \end{tabular}
    }
    \begin{tablenotes} 
        \footnotesize
        \item The best scores are marked in bold.
     \end{tablenotes}
    \end{threeparttable}
    \end{center}
	\vspace{-0.5cm}
    \end{table}
Another ablation is performed on the choices of feature maps for overlap estimation.
As shown in Tab.~\ref{table:ablation_overlap}, the performance, including IOU, precision, and recall consistently increases as the size of feature maps becomes smaller.
As mentioned in Section~\ref{method:overlap}, the deepest feature maps are the best scale to make overlap region classification.


\section{CONCLUSION}
We have presented a unified BEV model that jointly learns 3D local features and overlap estimation for point cloud registration and loop closure. 
The BEV representation makes it convenient to use a shared backbone for related multi-task processes. 
Overlap estimation plays a core role in significantly enhancing performance on both registration and loop closure, especially in low overlap scenarios. 
As a further extension of this work, we plan to add a new task head to generate a global descriptor that makes the method capable of place recognition at global scope retrieval. 
Furthermore, we will explore an end-to-end registration process that directly generates the relative transformation without RANSAC post-processing.

\bibliographystyle{ieeetr}
\bibliography{main}

\end{document}